\documentclass[conference]{IEEEtran}
\IEEEoverridecommandlockouts
\usepackage{cite}
\usepackage{amsmath,amssymb,amsfonts}
\usepackage{algorithmic}
\usepackage{graphicx}
\usepackage{textcomp}
\usepackage{xcolor}
\usepackage{hyperref}
\def\BibTeX{{\rm B\kern-.05em{\sc i\kern-.025em b}\kern-.08em
    T\kern-.1667em\lower.7ex\hbox{E}\kern-.125emX}}
\begin{document}

\title{Data Generation for Learning to Grasp in a Bin-picking Scenario\\

}

\author{\IEEEauthorblockN{Yiting Chen}
\IEEEauthorblockA{\textit{School of Power and Mechanical Engineering} \\
\textit{Wuhan University}\\
China \\
chenyiting@whu.edu.cn}
\and
\IEEEauthorblockN{Miao Li}
\IEEEauthorblockA{\textit{School of Power and Mechanical Engineering} \\
\textit{Wuhan University}\\
China \\
limiao712@gmail.com}
}

\maketitle

\begin{abstract}
The rise of deep learning has greatly transformed the pipeline of robotic grasping from model-based approach to data-driven stream. Along this line,  a large scale of grasping data either collected from simulation or from real world examples become extremely important. In this paper, we present our recent work on data generation in simulation for a bin-picking  scene. 77 objects from the YCB object data sets are used to generate the dataset with PyBullet, where different environment conditions are taken into account including lighting, camera pose, sensor noise and so on. In all, 100K data samples are collected in terms of ground truth segmentation, RGB, 6D pose and point cloud. All the data examples including the source code are made available online. 
\end{abstract}

\section{Introduction}
If we take a short look at recent years pose estimation and object location methods, data driven takes up an increasing proportion, such as CullNet\cite{9022530}, DenseFusion \cite{8953386}. These methods reveal reliable ways to estimate the 6D pose of objects, and of course out there are still many examples like this. With the help of large scale of data, the time to learn pose estimation or grasping has been significantly shortened.

\section{Building Dataset}

\subsection{System Setup}

\begin{figure}[htbp]
\centerline{\includegraphics{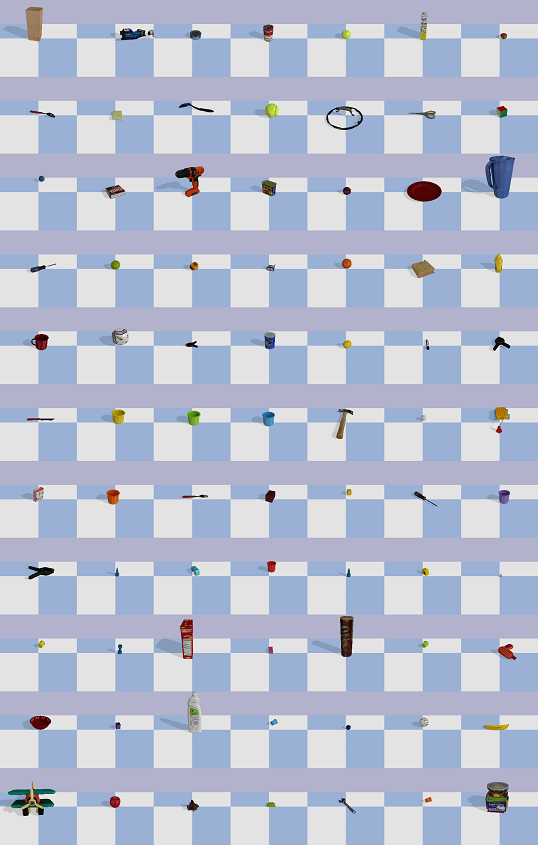}}
\caption{77 different kinds of objects.}
\label{pic1}
\end{figure}

\begin{itemize}
\item Simulation:We choose PyBullet as our simulator, which provides real-time collision detection and multi-physics simulation for VR, games, visual effects, robotics, machine learning etc..
\item Objects: All of our objects are selected from YCB-Dataset\cite{7251504}, which provides nearly hundred kinds of texture-mapped 3D mesh models. Figure \ref{pic1} shows 77 different kinds of objects.
\end{itemize}
\subsection{Data Generation}
\begin{figure}[htbp]
\centerline{\includegraphics{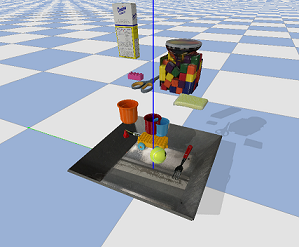}}
\caption{12 different objects appear randomly in the blank space above.}
\label{pic2}
\end{figure}

The virtual environment we designed is to place an empty tray box in the middle of a blank plane and the camera 0.7 meters above the tray box. There are 77 different kinds of models in our dataset, which are all selected from YCB dataset. We set a blank space of 0.4*0.4*0.45 cubic meters, and make it 0.05 meters right above the tray box. Each time we randomly selected 12 different kinds of models to appear from random positions above the box, every single object's x,y,z parameters were generated randomly within the size of the blank space. Figure \ref{pic2} shows the situation when 12 objects came out, which are sugar-box, g-cup, mug, sponge, a-colored-wood-blocks, c-lego-duplo, g-lego-duplo, scissors, large-marker, fork, h-cups, tennis-ball.

\begin{figure}[htbp]
\centerline{\includegraphics{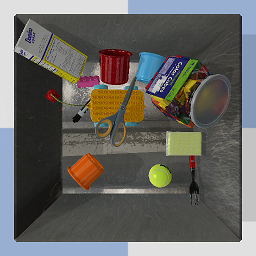}}
\caption{12 objects fall from above and become stable after collision.}
\label{pic3}
\end{figure}

As soon as we turned on gravity, the objects would naturally fall into the tray box. Due to the collision, the poses of each objects were naturally randomly generated, so that the stacking states of objects were very similar to the real world situation. Figure \ref{pic3} shows the situation after falling. For each falling case, the lighting of the scene comes from a point light that will constantly change its angle, which means we could obtain nearly every lighting situation that is possible in the real world.

\subsection{Simulation Result}
\begin{figure}[htbp]
\centerline{\includegraphics{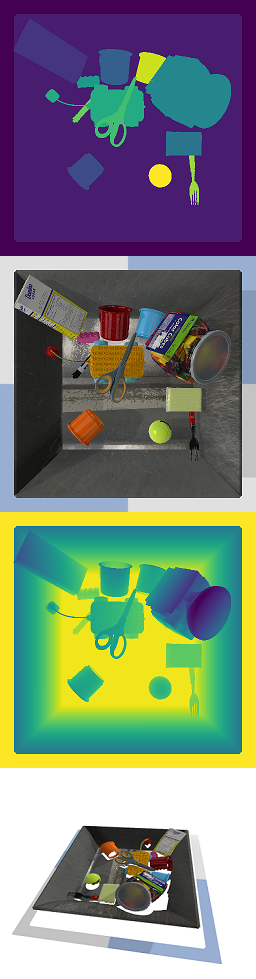}}
\caption{Segmentation, RGB, Depth and Point Cloud from top to bottom.}
\label{pic4}
\end{figure}

\begin{figure}[htbp]
\centerline{\includegraphics{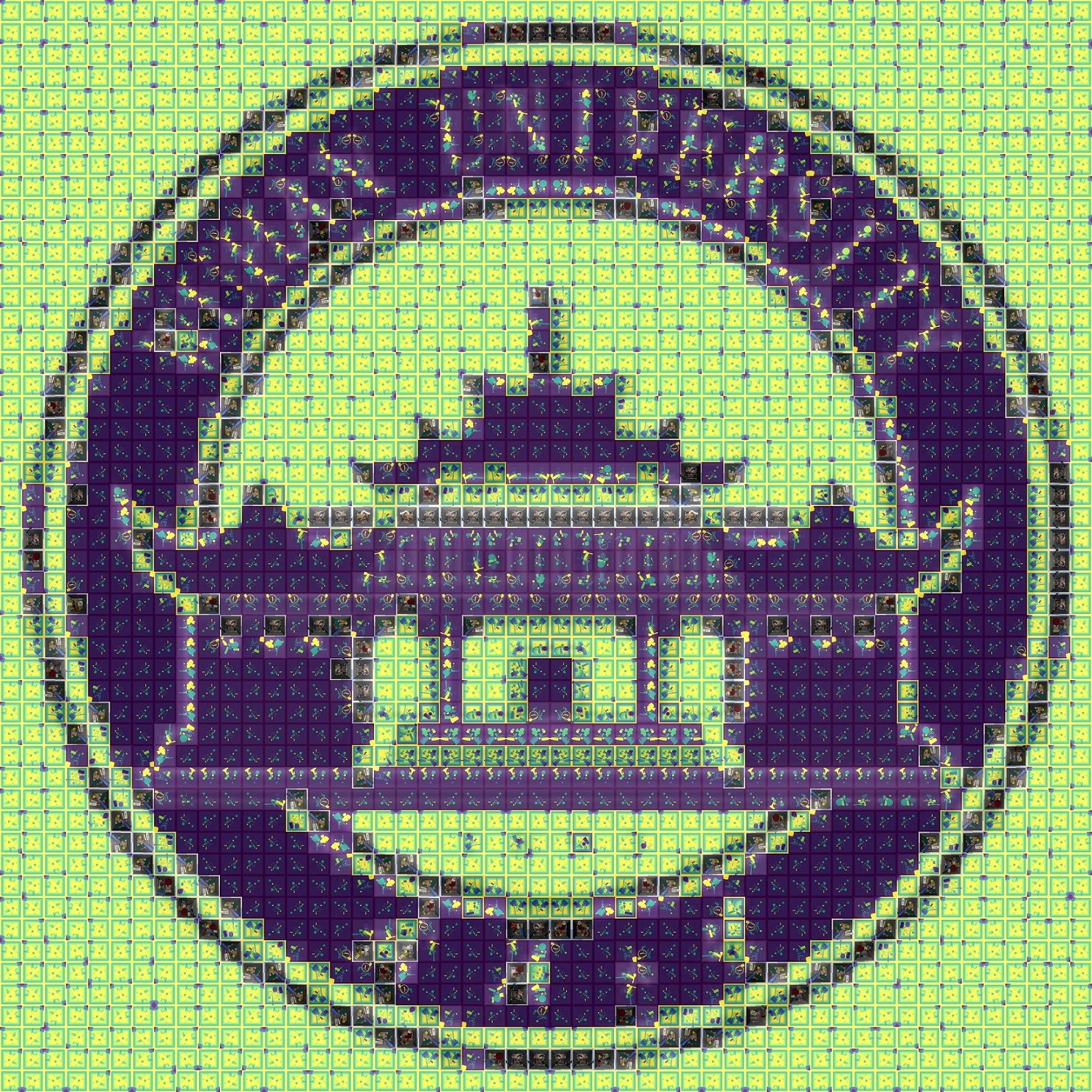}}
\caption{Part of our result, which contains images from 1200 groups of data}
\label{pic5}
\end{figure}

Thanks to the powerful build-in function from PyBullet, we could easily get segmentation, depth and RGB images of our tray box. Figure \ref{pic4} shows the 3 kinds of images and point cloud we get.All images are saved as .png file, point cloud is saved as .ply file. 
 Figure \ref{pic5} show cases part of our simulation result. 6D Poses of each object falling case are saved as .csv file, we describe the 6D Poses by quaternion.

\section{Conclusion}
We present a new dataset with point cloud, 6D pose ,segmentation,depth and RGB created using the PyBullet. This dataset includes 77 kinds of YCB models and includes random collision, lighting variations. Our Dataset contains 100k groups of data and provides significantly lots of parameter variations. In the future, we are planning to validate the effectiveness of this dataset using real world object examples.The website for the data generation procedure is available online as \url{cheneating716.github.io}

\bibliographystyle{./bibliography/IEEEtran}
\bibliography{./bibliography/IEEEabrv,./bibliography/IEEEexample}

\end{document}